\documentclass[11pt]{article}

\usepackage[final]{acl}

\usepackage{times}
\usepackage{latexsym}

\usepackage[T1]{fontenc}

\usepackage[utf8]{inputenc}

\usepackage{microtype}
\usepackage{inconsolata}

\usepackage{graphicx}
\usepackage{placeins}

\usepackage{amsmath}
\usepackage{amssymb}

\usepackage{booktabs}
\usepackage{multirow}

\usepackage{xcolor}
\newif\ifchg \chgfalse
\newcommand{\chg}[1]{\ifchg\textcolor{red}{#1}\else#1\fi}

\usepackage{tikz}
\usetikzlibrary{positioning,arrows.meta,fit,calc,backgrounds}


\title{GRAFT: Cross-Backbone LoRA Reuse for Continual Language and Multimodal Learning}

\author{
  Ha Pham Thien Dinh \\
  Deakin University \\
  \texttt{s.dinh@deakin.edu.au} \And
  Ming Liu \\
  Deakin University \\
  \texttt{m.liu@deakin.edu.au} \And
  Jiesi Guo \\
  Australian Catholic University \\
  \texttt{jiesi.guo@acu.edu.au} \AND
  Matteo Vergani \\
  Deakin University \\
  \texttt{matteo.vergani@deakin.edu.au} \And
  Kewen Liao \\
  Deakin University \\
  \texttt{kewen.liao@deakin.edu.au}
}

\begin{document}

\maketitle


\begin{abstract}
In continual learning, parameter-efficient fine-tuning enables large language models to acquire new capabilities through lightweight adapters such as LoRA, yet the reuse of pretrained adapters is typically restricted to models sharing the same backbone. This limits the transfer of learned capabilities from language models to multimodal models. We propose \textbf{GRAFT} (Generalist Reuse of Adapters via Feature-space Transfer), a framework for cross-backbone LoRA reuse. GRAFT retrieves the most relevant pretrained LoRA adapter from a modality-aware library and maps it into the target backbone through a learned linear bridge between corresponding source and target feature spaces. We evaluate GRAFT on ten Super-NaturalInstructions text tasks, transferring Qwen2.5 LoRAs to Qwen2.5-VL, and five multimodal tasks, where Qwen2.5-VL LoRAs are directly reused. Compared with randomly initialized LoRA adapters, GRAFT improves performance on most evaluated tasks. These results show that GRAFT provides a simple and effective mechanism for reusing learned capabilities across language and multimodal models in continual learning.

\end{abstract}

\section{Introduction}
Large language models (LLMs) and vision-language models (VLMs) increasingly need to acquire new capabilities over time as tasks, domains, and modalities evolve \cite{overall_survey}. Parameter-efficient fine-tuning (PEFT), particularly Low-Rank Adaptation (LoRA) \cite{lora_base}, provides an attractive mechanism for continual learning by adapting only a small number of parameters while keeping the backbone frozen. New capabilities can therefore be stored as lightweight adapters rather than separate copies of the full model, reducing both training and deployment costs. The growing availability of pretrained LoRA adapters creates an additional opportunity for continual learning: instead of learning each new adapter from random initialization, previously trained adapters can provide useful priors for subsequent tasks. AdaEvol \cite{adaevol} introduced this idea on LLMs by retrieving relevant pretrained LoRAs for new-task adaptation, showing that knowledge encoded in existing adapters can remain useful after further supervised fine-tuning. However, such reuse is typically restricted to adapters trained on the same LLM backbone. This substantially limits the knowledge available to a continual multimodal learner (e.g., VLMs) when relevant capabilities have been learned by a closely related but different model (e.g., LLMs).

We study this problem using Qwen2.5-7B-Instruct \cite{qwen25} as a source language model and Qwen2.5-VL-7B-Instruct \cite{qwen25vl} as the target multimodal model. The two models belong to the same family and contain closely related language components, making them a natural testbed for cross-backbone adapter reuse. Many text-side capabilities, including reasoning, question answering, and domain-specific knowledge, may already be encoded in Qwen2.5 LoRAs even when equivalent Qwen2.5-VL adapters are unavailable. Reusing these adapters would allow a multimodal continual learner to access a broader pool of pretrained capabilities while consolidating them into a single Qwen2.5-VL backbone.

Cross-backbone reuse is nevertheless non-trivial. A LoRA adapter is optimized in the feature space induced by its source backbone, and directly attaching it to a different backbone does not preserve the activation environment in which its parameters were learned \cite{translora}. We therefore propose \textbf{GRAFT} (Generalist Reuse of Adapters via Feature-space Transfer), a cross-backbone warm-start framework for multimodal continual learning. GRAFT retrieves a relevant pretrained LoRA from a modality-aware LoRA library and estimates a lightweight linear compatibility bridge between corresponding source and target feature spaces. This enables a generalist continual learner to reuse both text-only LoRAs from Qwen2.5 and multimodal LoRAs from Qwen2.5-VL. For cross-backbone tasks, retrieved Qwen2.5 LoRAs are transferred through GRAFT and fine-tuned into Qwen2.5-VL-compatible adapters; for same-backbone tasks, Qwen2.5-VL LoRAs are reused directly. In both cases, the resulting LoRA adapters are retained for future retrieval, allowing the library to grow over time while deployment remains centered on a single multimodal backbone. Our main contributions are as follows:

\begin{itemize}

\item We introduce GRAFT, a single-backbone multimodal continual learning framework that reuses both text-only and multimodal LoRA libraries through modality-aware retrieval, enabling a Qwen2.5-VL learner to access a broader pool of previously acquired capabilities while keeping deployment centered on a single multimodal backbone.

\item We propose a lightweight cross-backbone warm-start mechanism in GRAFT that learns an activation-space compatibility bridge between Qwen2.5 and Qwen2.5-VL and applies it to the retrieved LoRA adapter's A matrix. The bridge is folded into the adapted LoRA after training, yielding a standard Qwen2.5-VL adapter with no inference-time overhead.

\item We evaluate cross-backbone adapter reuse on ten Super-NaturalInstructions text tasks and same-backbone reuse on five multimodal tasks. GRAFT consistently outperforms adaptation from randomly initialized Qwen2.5-VL LoRAs, demonstrating its superiority of adapter reuse for continual capability acquisition across language and multimodal settings.

\end{itemize}

\section{Related Work}

\subsection{Parameter-Efficient Continual Learning}

LoRA represents a task update as a low-rank correction to frozen pretrained
weights, so adapters can be stored and exchanged without a full copy of the
model \cite{lora_base}. Later variants refine how that budget is
spent \cite{adalora,vera,dora} or reduce the cost of training them
\cite{qlora,lora_plus,lora_fa}. Closest to our own concern is PiSSA
\cite{pissa}, which builds the trainable matrices from the pretrained weight's
principal singular directions rather than at random, showing that a stronger
starting point already exists in the backbone.

Continual-learning methods also isolate task-specific changes to protect
earlier capabilities, through a separate adapter or prompt per task
\cite{adaptercl,progressive_prompts}, approximately orthogonal low-rank updates
\cite{olora}, or blocks coordinated by shared attention \cite{sapt}. These
works treat adapters as
persistent, reusable units of knowledge, but construct and reuse them around a
single backbone.

\subsection{Adapter Selection, Routing, and Composition}
A second line treats trained adapters as a pool to select from or compose. AdapterFusion trains task-specific adapters independently, then learns how their representations contribute to a target task \cite{adapterfusion}, and LoraHub extends this to LoRA libraries by optimizing combination weights from few examples \cite{lorahub}. Dynamic variants gate those contributions or treat LoRAs as hierarchically gated experts \cite{loraflow,mole}, others route among adapters as tasks arrive \cite{l2r}, and AdaEvol retrieves pretrained LoRAs to initialise adaptation on the same backbone \cite{adaevol}. All of these assume a shared backbone; we ask how such a pool can span two.


\subsection{Cross-Model Adapter Transfer}

Closest to our setting are works that transfer adapters across model backbones. Trans-LoRA distills a source LoRA into a newly initialized target adapter using synthetic target-task data filtered by a discriminator \cite{translora}. LoRA-X instead constrains adapters during training to the source backbone's singular subspace and projects them into compatible target subspaces \cite{lorax}. \chg{Following either method requires inputs our setting does not provide. LoRA-X transfers only adapters trained under its own subspace constraint; imposing that constraint on a retrieved adapter would mean retraining it on the corpus it was originally built from, and several of those corpora were never released. Trans-LoRA instead substitutes synthetic data for target-task data where none is available, whereas each of our tasks supplies its own. We therefore use an ordinary pretrained LoRA as an initialization prior.}



\section{Methodology}

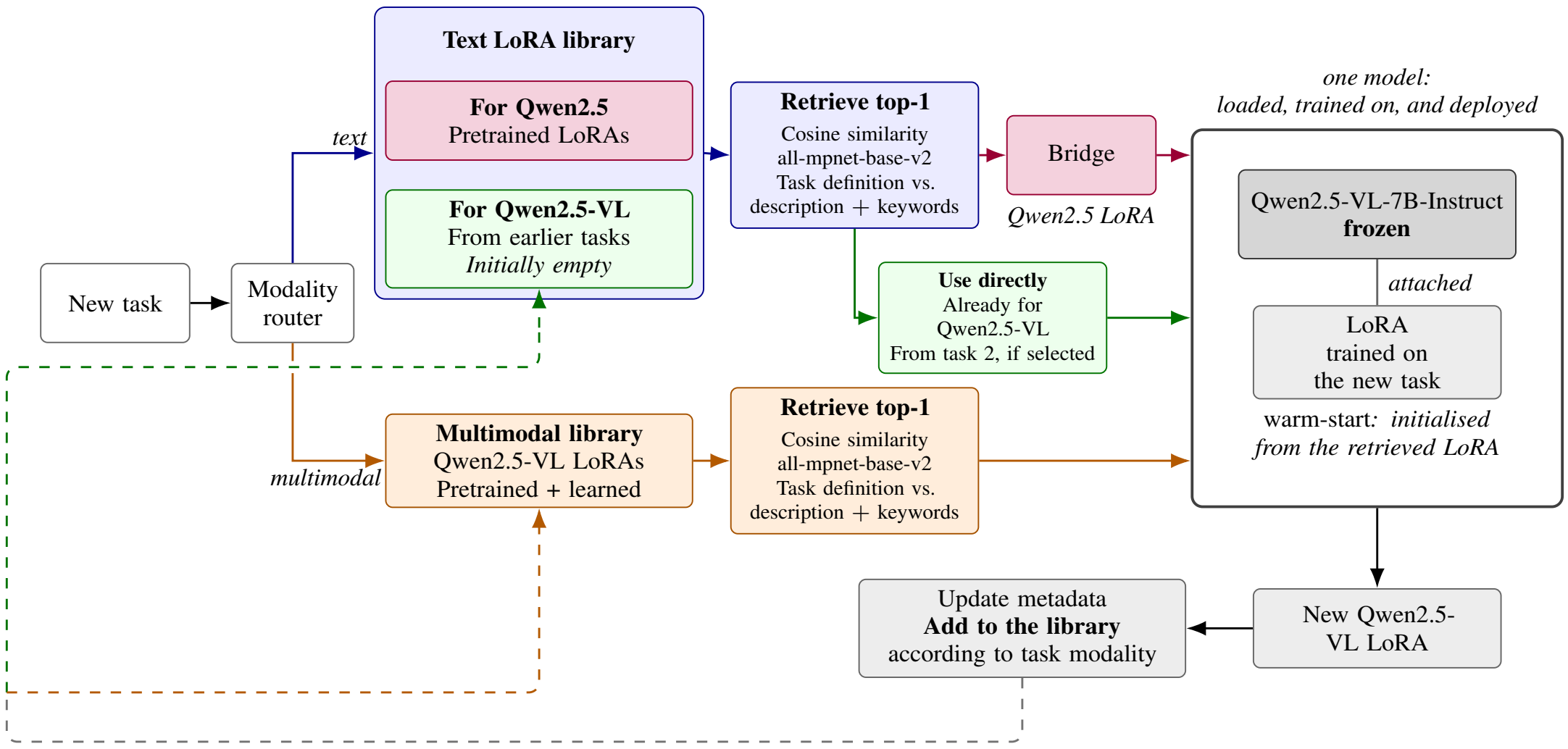
\begin{figure*}[!t]
\centering
\scalebox{0.90}{%
\begin{tikzpicture}[
  blk/.style={draw, rounded corners=2pt, align=center, font=\scriptsize,
              inner sep=3pt, minimum height=8mm, text width=16mm},
  txt/.style={blk, fill=blue!8,    draw=blue!55!black, text width=14mm},
  txtp/.style={txt, text width=25mm},
  brg/.style={blk, fill=purple!18, draw=purple!80!black, text width=13mm},
  sourceblk/.style={blk, fill=purple!8, draw=purple!80!black},
  directblk/.style={blk, fill=green!7, draw=green!45!black},
  vis/.style={blk, fill=orange!14, draw=orange!65!black, text width=14mm},
  visp/.style={vis, text width=25mm},
  shd/.style={blk, fill=black!7,   draw=black!60, text width=23mm},
  ent/.style={blk, fill=white,     draw=black!60},
  bbx/.style={draw=black!75, thick, rounded corners=3pt, inner sep=4mm},
  llm/.style={blk, fill=black!16, draw=black!80, text width=26mm, minimum height=9mm},
  ar/.style={-Latex, semithick},
  arb/.style={ar, draw=blue!55!black},
  arbridge/.style={ar, draw=purple!80!black},
  arg/.style={ar, draw=green!45!black},
  arv/.style={ar, draw=orange!70!black},
  fb/.style={-Latex, semithick, dashed, draw=black!55},
  tag/.style={font=\scriptsize\itshape, inner sep=2pt}
]

\node[ent, text width=13mm] (task)   at (0,0)   {New task};
\node[ent, text width=11mm, inner sep=2pt] (router) at (1.8,0)
      {Modality\\router};

\node[font=\scriptsize\bfseries] (texttitle) at (4.3,2.65)
      {Text LoRA library};
\node[brg, text width=29mm] (textpretrained) at (4.3,1.85)
      {\textbf{For Qwen2.5}\\Pretrained LoRAs};
\node[directblk, text width=29mm]
      (textlearned) at (4.3,0.65)
      {\textbf{For Qwen2.5-VL}\\From earlier tasks\\\emph{Initially empty}};
\begin{scope}[on background layer]
\node[fill=blue!8, draw=blue!55!black, rounded corners=3pt, inner sep=3pt,
      fit=(texttitle)(textpretrained)(textlearned)] (ptext) {};
\end{scope}
\node[visp, text width=29mm] (pvl) at (4.3,-1.6)
      {\textbf{Multimodal library}\\Qwen2.5-VL LoRAs\\Pretrained + learned};

\node[txt, text width=23mm, font=\tiny] (rtext) at (7.5,1.5)
      {{\scriptsize\textbf{Retrieve top-1}}\\[2pt]
       Cosine similarity\\all-mpnet-base-v2\\
       Task definition vs.\\description $+$ keywords};
\node[vis, text width=23mm, font=\tiny] (rvl) at (7.5,-1.6)
      {{\scriptsize\textbf{Retrieve top-1}}\\[2pt]
       Cosine similarity\\all-mpnet-base-v2\\
       Task definition vs.\\description $+$ keywords};

\node[brg] (bridge) at (9.8,1.5) {Bridge};
\node[tag, below=1pt of bridge] {Qwen2.5 LoRA};
\node[directblk, text width=21mm,
      font=\tiny] (direct) at (8.9,-0.15)
      {\textbf{Use directly}\\Already for Qwen2.5-VL\\From task 2, if selected};

\node[llm] (llm)  at (12.8, 0.90) {Qwen2.5-VL-7B-Instruct\\\textbf{frozen}};
\node[shd] (warm) at (12.8,-0.50) {LoRA\\trained on the new task};
\node[tag, below=1pt of warm, align=center, text width=28mm] (wsnote)
     {\emph{warm-start}: initialised\\from the retrieved LoRA};
\draw[semithick, draw=black!60] (llm.south) -- node[tag, right=1pt] {attached} (warm.north);
\node[bbx, fit=(llm)(warm)(wsnote)] (backbone) {};
\node[tag, above=1pt of backbone, align=center]
     {one model:\\loaded, trained on, and deployed};

\node[shd] (out) at (12.8,-3.3) {New Qwen2.5-VL LoRA};
\node[shd, text width=31mm] (addlibrary) at (9.2,-3.3)
      {Update metadata\\\textbf{Add to the library}\\according to task modality};

\draw[ar]  (task) -- (router);
\draw[arb] (router.north) |- node[tag, above, pos=0.85] {text} (ptext.west);
\draw[arv] (router.south) |- node[tag, below, pos=0.68] {multimodal} (pvl.west);
\draw[arb] (ptext.east) -- (rtext.west);
\draw[arv] (pvl)   -- (rvl);
\draw[arbridge] (rtext) -- (bridge);

\draw[arbridge] (bridge.east) -- (backbone.west |- bridge.east);
\draw[arg] (rtext.south) |- (direct.west);
\draw[arg] (direct.east) -- (backbone.west |- direct.east);
\draw[arv] (rvl.east) -- (backbone.west |- rvl.east);

\draw[ar]  (backbone.south) -- (out.north);

\draw[ar] (out.west) -- (addlibrary.east);
\coordinate (libraryfork) at (-1.1,-3.95);
\draw[semithick, dashed, draw=black!55, rounded corners=3pt]
      (addlibrary.south) -- (9.2,-4.45) -- (-1.1,-4.45) -- (libraryfork);
\draw[semithick, dashed, draw=green!45!black, rounded corners=3pt,
      preaction={draw=white, line width=3.5pt}]
      (libraryfork) -- (-1.1,-0.65) -- (4.3,-0.65) -- (4.3,-0.25);
\draw[fb, draw=green!45!black]
      (4.3,-0.25) -- (textlearned.south);
\draw[fb, draw=orange!70!black, rounded corners=3pt]
      (libraryfork) -- (4.3,-3.95) -- (pvl.south);

\end{tikzpicture}%
}
\caption{GRAFT pipeline. A task retrieves one LoRA from its modality's
library. The text library separates pretrained Qwen2.5 LoRAs
(purple, bridge required) from Qwen2.5-VL LoRAs learned on earlier tasks
(green, used directly). The latter group is initially empty and can be used
from task 2 onward, if selected.
Multimodal LoRAs already work with Qwen2.5-VL. After training, metadata is
updated and the LoRA is added to the appropriate library. Solid arrows show
task processing; dashed arrows show library additions. The bridge is detailed
in Figure~\ref{fig:bridge}.}

\label{fig:pipeline}
\end{figure*}

In this section, we describe and formulate GRAFT, a generalist continual-learning framework that retrieves reusable LoRAs from text and multimodal adapter libraries and
consolidates their capabilities into Qwen2.5-VL-7B-Instruct. When the
retrieved adapter originates from Qwen2.5-7B-Instruct, GRAFT applies a
cross-backbone warm-start mechanism based on an activation-space bridge;
adapters already trained for Qwen2.5-VL are used directly. The bridged
adapter is then optimized on the incoming task, after which the bridge is
folded into the LoRA $A$ matrix to produce a standard Qwen2.5-VL LoRA.
Figure~\ref{fig:pipeline} gives an overview of the complete framework.

\subsection{Problem Formulation}

We consider a sequence of tasks $\{\mathcal{T}_1,\ldots,\mathcal{T}_K\}$
arriving over time, where each task $\mathcal{T}_t$ provides training data
$\mathcal{D}_t$. The target model is Qwen2.5-VL-7B-Instruct, denoted by $M_T$,
whose backbone remains frozen during parameter-efficient adaptation. For a
target weight $W_T$, LoRA introduces a trainable low-rank update

\begin{equation}
    W_T' = W_T + \Delta W,
    \qquad
    \Delta W = BA,
\end{equation}

where $A \in \mathbb{R}^{r \times d_{\mathrm{in}}}$ and
$B \in \mathbb{R}^{d_{\mathrm{out}} \times r}$ are the trainable LoRA
parameters.

Alongside adapters previously trained on $M_T$, we consider pretrained LoRAs
originating from the related Qwen2.5-7B-Instruct source model $M_S$. Such an
adapter $\Delta W_S$ was trained inside $M_S$, and everything it learned was
learned against the activations that model produces.

That is what makes reuse across the two backbones non-trivial. A LoRA's
contribution to a module's output, $x A^{\top} B^{\top}$, is a function of the
activations $x$ that the module receives, and those activations belong to the
backbone rather than to the adapter. Continued multimodal training has moved
them, so the same adapter attached to $M_T$ reads inputs unlike any it was
fitted on, and the correction it applies is no longer the correction it
learned. We therefore locate cross-backbone compatibility in the feature
spaces the two models produce, and build the bridge there: for each module we
estimate, by least squares on paired activations, the linear map that carries
target-backbone features back into the source backbone's feature space.

Our objective is not to make the source LoRA directly transferable without
additional training. Given $\Delta W_S$ and the incoming task data
$\mathcal{D}_t$, we use the pretrained adapter as an informative
initialization and optimise it into a LoRA $\Delta W_t^{*}$ for Qwen2.5-VL, so
that the final adapter keeps the benefit of pretrained knowledge while
operating entirely on the Qwen2.5-VL backbone.

\subsection{Activation-Space Compatibility Bridge}

\begin{figure*}[!t]
\centering
\scalebox{0.90}{%
\begin{tikzpicture}[
  blk/.style={draw, rounded corners=2pt, align=center, font=\scriptsize,
              inner sep=3pt, minimum height=8mm, text width=16mm},
  brg/.style={blk, fill=purple!18, draw=purple!80!black, text width=13mm},
  sourceblk/.style={blk, fill=purple!8, draw=purple!80!black},
  shd/.style={blk, fill=black!7, draw=black!60, text width=23mm},
  ar/.style={-Latex, semithick},
  arbridge/.style={ar, draw=purple!80!black}
]
\draw[draw=purple!80!black, thick, rounded corners=3pt]
      (-0.8,3.55) rectangle (14.65,7.95);
\node[font=\small\bfseries, text=purple!80!black] at (6.925,7.62)
      {Inside the bridge: let a source LoRA read target-model features};
\draw[draw=purple!20] (5.45,3.8) -- (5.45,7.25);

\node[font=\scriptsize\bfseries] at (2.25,7.12)
      {1. Fit once offline};
\node[font=\scriptsize] (calibtokens) at (2.25,6.72)
      {Same calibration tokens};
\node[sourceblk, text width=21mm] (calibsource) at (0.65,6.05)
      {Qwen2.5\\frozen source};
\node[shd, text width=21mm] (calibtarget) at (3.85,6.05)
      {Qwen2.5-VL\\frozen target};
\draw[ar] (calibtokens.west) -| (calibsource.north);
\draw[ar] (calibtokens.east) -| (calibtarget.north);
\node[font=\scriptsize, align=center] (sourcefeatures) at (0.65,5.32)
      {$X_S$: source features};
\node[font=\scriptsize, align=center] (targetfeatures) at (3.85,5.32)
      {$X_T$: target features};
\draw[arbridge, shorten >=1pt] (calibsource.south) -- (sourcefeatures.north);
\draw[ar, shorten >=1pt] (calibtarget.south) -- (targetfeatures.north);
\node[brg, text width=43mm, minimum height=9mm] (bridgefit) at (2.25,4.37)
      {Fit $T$ by ridge regression\\$X_T T \approx X_S$};
\draw[arbridge] (sourcefeatures.south) -- (sourcefeatures.south |- bridgefit.north);
\draw[arbridge] (targetfeatures.south) -- (targetfeatures.south |- bridgefit.north);
\node[font=\scriptsize, text=black!70, align=center, text width=57mm] at (2.25,3.79)
      {One map per module input space; reused across tasks};

\node[font=\scriptsize\bfseries] at (10.0,7.12)
      {2. Train the retrieved LoRA on the target task};
\node[font=\scriptsize, align=center, text width=12mm] (detailinput) at (6.25,6.13)
      {$x_T$\\target\\features};
\node[brg, text width=12mm] (detailmap) at (7.85,6.13)
      {$T$\\fixed};
\node[sourceblk, text width=13mm] (detaila) at (9.7,6.13)
      {$A_t^{\top}$\\trainable};
\node[sourceblk, text width=13mm] (detailb) at (11.55,6.13)
      {$B_t^{\top}$\\trainable};
\node[font=\scriptsize, align=center] (detailoutput) at (13.55,6.13)
      {$\Delta y$\\LoRA output};
\draw[arbridge] (bridgefit.east) -- (5.15,4.37) -- (5.15,6.75)
      -- node[font=\scriptsize, above, fill=white, inner sep=1pt] {reuse $T$}
      (7.85,6.75) -- (detailmap.north);
\draw[arbridge] (detailinput) -- (detailmap);
\draw[arbridge] (detailmap) -- (detaila);
\draw[arbridge] (detaila) -- (detailb);
\draw[arbridge] (detailb) -- node[font=\scriptsize, above] {$\times s$} (detailoutput);
\node[font=\scriptsize, align=center, text width=83mm] at (10.0,5.32)
      {$\Delta y = s\,x_T T A_t^{\top} B_t^{\top}$\\
       Target features are mapped into the source feature space.};

\node[shd, text width=78mm, minimum height=10mm] (bridgefold) at (10.0,4.3)
      {\textbf{3. Fold for deployment}\\
       $A^{*}=A_t T^{\top},\quad B^{*}=B_t$\\
       Standard target LoRA: same matrix shapes, no bridge at inference};
\end{tikzpicture}%
}
\caption{Activation-space bridge. Paired activations from the frozen backbones
define $T$ once offline. During target-task adaptation, $T$ stays fixed while
the LoRA $A$ and $B$ matrices train. Folding $T$ into the trained $A$ matrix produces a Qwen2.5-VL LoRA with
no separate bridge at inference. Only offline calibration needs the source model.}
\label{fig:bridge}
\end{figure*}
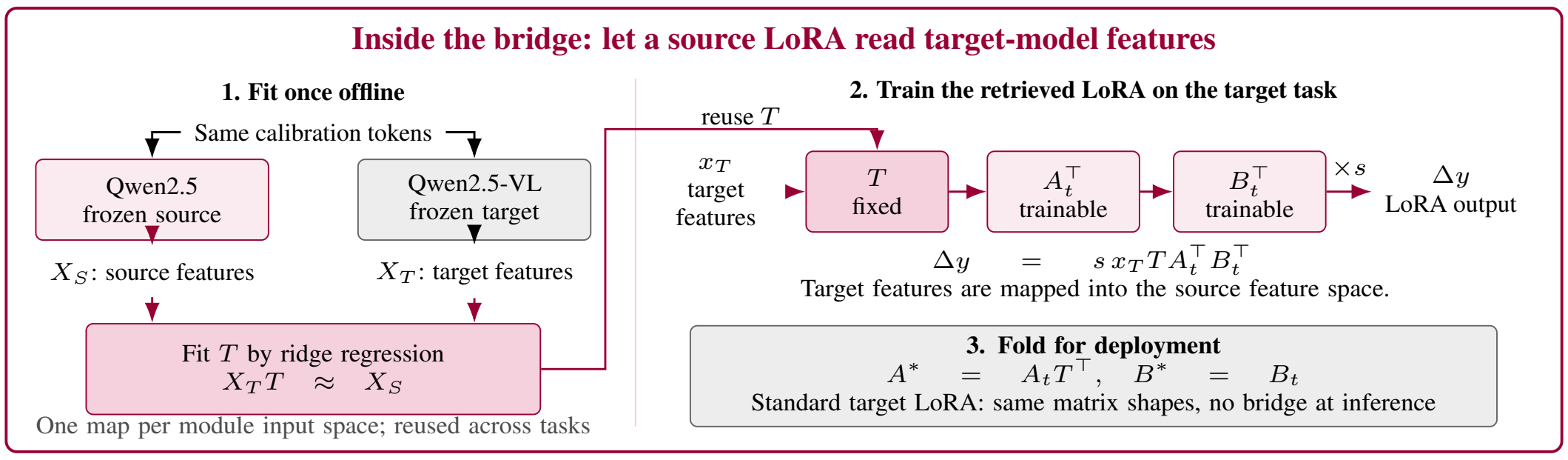

Qwen2.5 and Qwen2.5-VL belong to the same model family and contain closely
corresponding language layers, so a module's input activations in the two
backbones are in principle comparable. Figure~\ref{fig:bridge} illustrates
offline bridge fitting, LoRA training with the bridge fixed, and folding for
deployment. We seek a linear map that
carries target-backbone features back into the source backbone's feature
space, so that the source adapter's read-out map remains functionally valid.

Let $X_S, X_T \in \mathbb{R}^{n \times d_{\mathrm{in}}}$ collect the input
activations of a given module when the \emph{same} token sequences are pushed
through $M_S$ and $M_T$ respectively. We seek a transfer matrix $T$ under which
the adapter's output change is preserved,

\begin{equation}
    X_T\,T \; \approx \; X_S ,
\end{equation}

which we obtain in closed form as a ridge regression with coefficient
$\lambda$,

\begin{equation}
    T
    =
    \bigl(X_T^{\top} X_T + \lambda I\bigr)^{-1}
    X_T^{\top} X_S .
\end{equation}
\chg{$T$ is computed offline from a separate dataset, external to the task
sequence, that contains no test item;} details are given in
Appendix~\ref{app:bridge-calibration}.

To approximate $X_S \Delta W_S^{\top}$ on target features, we use
$\Delta W_T = \Delta W_S T^{\top}$. Substituting the factored form
$\Delta W = BA$ shows that the correction applies to the LoRA $A$ matrix:

\begin{equation}
    A' = A T^{\top},
    \qquad
    B' = B,
    \qquad
    s' = s ,
\end{equation}

where $s$ is the adapter's scaling factor. The LoRA $B$ matrix and the scaling
are left unchanged, while the rank budget is preserved and the update remains
represented by two low-rank matrices. The correction is applied only to the LoRA $A$ matrix because
$A$ is the matrix that reads the backbone activations, where the discrepancy
between the two models arises.

The activation map addresses a different question from a correction based on
backbone weight differences. Weight differences can identify where the models
have changed, and can be used to discount an update or project out selected
directions. They do not directly estimate how the features presented to an
adapter correspond across models. The bridge instead fits that correspondence
on paired activations: it can mix input coordinates so that the source
adapter reads a transformed version of the target features. Its objective is
to recover the source read-out behaviour before task-specific optimisation,
not simply to reduce the magnitude of a transplanted update. This remains an
approximation whose quality depends on how well a linear map describes the
two feature spaces; fitting a square matrix does not guarantee invertibility
or exact preservation of the source model's behaviour.

Within a decoder layer several projections read the same tensor: the query,
key and value projections share one input, and the two MLP input projections
share another. There are therefore fewer distinct input spaces than target
modules, and a single $T$ is solved per space and shared by the modules that
read it. We omit the MLP output projection, whose input dimension makes its
transfer matrix disproportionately expensive to store, so the bridge covers
six of the seven LoRA-target projections; modules without a transfer matrix
pass through unchanged.

The transfer matrix depends only on the two backbones and the calibration data,
never on any particular adapter. It is therefore computed once, offline, and
reused for every source LoRA transferred between the same pair of models: it
represents model-level compatibility rather than a transformation learned per
adapter. Because the fit is a regression, its residual on held-out sequences
also serves as a direct, training-free test of whether the two feature spaces
are linearly relatable, which we assess by comparing the bridged map against the
uncorrected identity map.

\subsection{Warm-Start Adaptation}

For an incoming task, the retrieved Qwen2.5 LoRA supplies the initial
parameters $A_0$ and $B_0$. The transfer matrices are retained in the forward
computation throughout the single fine-tuning stage. At optimisation step $t$,
the module's LoRA contribution is
\begin{equation}
    x_T \longmapsto s\,x_T T A_t^{\top} B_t^{\top},
\end{equation}
where $x_T$ comes from Qwen2.5-VL. The backbone and $T$ are frozen, while both
$A_t$ and $B_t$ are trained on the new task. The bridge supplies the feature
mapping through which the retrieved matrices initially operate; supervision
then adapts their response to the target model. Warm-starting therefore names
where this training begins, rather than an additional training stage.

Keeping $T$ active also distinguishes this training parameterisation from
using $AT^{\top}$ as an initial value and then optimising that product freely.
Here, optimisation updates $A_t$ behind the same fixed map at every step.
The adapter can change its response to the incoming task, while the
correspondence estimated during calibration continues to define how target
features enter it. This does not require the trained LoRA matrices to remain close
to their source values or guarantee preservation of the original task.

Because $T$ enters only through the LoRA $A$ matrix, the effective update is
\begin{equation}
    \Delta W_{\mathrm{eff}} = s\,B_t\bigl(A_t T^{\top}\bigr).
\end{equation}
The consecutive linear maps can be composed without storing a dense weight
update. On completion of fine-tuning we fold the bridge into the LoRA $A$ matrix
and store
\begin{equation}
    A^{*} = A_t T^{\top}, \qquad B^{*} = B_t.
\end{equation}
Since $T$ is square, $A^{*}$ retains shape $r\times d_{\mathrm{in}}$ and
$B^{*}$ retains shape $d_{\mathrm{out}}\times r$. These are precisely the
matrix shapes required by a standard LoRA attached to the corresponding
Qwen2.5-VL module. Its output satisfies
$s\,x_T(A^{*})^{\top}(B^{*})^{\top}
=s\,x_T T A_t^{\top}B_t^{\top}$, so the saved adapter accepts target-model
activations directly and computes the same update as the trained bridged
adapter. The activation fit is approximate, but this folding step is
algebraically exact and does not require $T$ to be invertible.

This explains why the result is compatible with Qwen2.5-VL: its matrices have
been optimised on that backbone, and its stored LoRA $A$ matrix already includes
the feature transformation used during training. The source model and a
separate bridge are no longer needed at inference. Evaluation and any later
warm-start can load the saved matrices as an ordinary Qwen2.5-VL LoRA;
applying the bridge again would transform them twice. Each task keeps its own
adapter and the shared backbone remains frozen, so subsequent adaptation
cannot overwrite the parameters used to serve an earlier task.

\subsection{Generalist Adapter Library}

GRAFT maintains two libraries organised by task modality,
$\mathcal{L}=\mathcal{L}_{\mathrm{Text}}\cup\mathcal{L}_{\mathrm{VL}}$.
The public LoRAs used to initialise them vary in rank, scaling factor, dropout
and target-module selection. The bridge operates on the $A$ matrix an
adapter provides without changing its shape; it requires no common rank or
module configuration across the library. The collected adapters and their
configurations are described in the appendix.

The text library contains the two groups shown in Figure~\ref{fig:pipeline}.
The first holds pretrained LoRAs \emph{for Qwen2.5}; these need the bridge
when attached to Qwen2.5-VL. The second holds LoRAs \emph{for Qwen2.5-VL}
learned on earlier text tasks. This group is initially empty and can first
supply an adapter at task 2, provided an earlier text task has completed.
The multimodal library contains pretrained and newly learned Qwen2.5-VL
LoRAs, which can be used directly. Library membership thus identifies the
kind of task an adapter can be retrieved for, while its recorded backbone
compatibility determines whether it needs the bridge.

\chg{A modality router selects the appropriate library from a task's modality flag, or otherwise from whether its samples contain images. Routing is rule-based and deterministic.}
Within that library, a single retrieval step ranks all candidates. In
particular, the two groups in the text library are ranked together: a
Qwen2.5 adapter takes the bridge path, while a selected Qwen2.5-VL adapter
takes the \emph{Use directly} path. The latter is an option for later tasks,
not an additional source of adapters available at the start of the sequence.

Each public adapter carries a description and keywords generated once offline
by prompting \chg{Claude Opus 5} with its model card. A fixed schema records
the task, domain, inputs, outputs and capabilities. We encode the incoming
task definition and both metadata fields with \texttt{all-mpnet-base-v2},
score each adapter by the equally weighted cosine similarities to its
description and keywords, and retrieve the top-ranked LoRA. This requires
neither loading candidate weights nor inspecting test examples. No loss-based
reranking is used for the reported results.

After fine-tuning and any bridge folding, metadata is updated to describe the
new task, identify the initialising adapter and record compatibility with
Qwen2.5-VL. The trained LoRA is then \emph{added to the library} for its task's
modality. Text outputs enter the Qwen2.5-VL group, so any later retrieval uses
them without the bridge. Updating the task description makes the newly
learned capability visible to subsequent similarity searches. The library can
therefore grow across a sequence while every completed task remains served by
its own saved adapter.



\FloatBarrier
\section{Experiments}

\subsection{Datasets and Tasks}

We evaluate GRAFT on both text-only and multimodal tasks, with all experiments using Qwen2.5-VL-7B-Instruct as the frozen target backbone. For text evaluation, we use ten tasks from Super-NaturalInstructions \cite{sni}, covering entailment, sentiment analysis and verification, review rating prediction, emotion classification, and answerability. For multimodal evaluation, we use five tasks: four image classification tasks (aerial scene, fine-grained dog breed, document image and waste classification) and OCRVQA, an OCR-based visual question answering task. The tasks are not selected arbitrarily: each has a semantically related capability in our pretrained LoRA libraries, which hold 28 text and 22 multimodal adapters, so the evaluation tests cross-backbone warm-starting under the intended retrieval scenario. Full task and library details are given in the appendix. \chg{The bridge is computed offline from the task set of \cite{adaevol}, before the task sequence begins.}

Because the library grows, task order determines which earlier adapters are
available for retrieval. We evaluate four orders fixed before running:
text first, multimodal first, interleaved, and fully shuffled. Only the last
changes the within-modality sequence. Appendix~\ref{sec:orders} lists
the sequences, and Table~\ref{tab:app-perorder} reports their results.

\subsection{Baselines}

\textit{LoRA-Random} is compared against every warm-started configuration
using the same target-task data and number of optimisation steps in both
settings: it
fine-tunes a newly initialised Qwen2.5-VL LoRA on each task.
\textit{GRAFT} instead initialises that adapter from a retrieved Qwen2.5 LoRA, keeps the bridge fixed during training,
and folds it into the saved adapter.
Because the multimodal library already holds Qwen2.5-VL adapters, a retrieved
LoRA needs no bridge there; we call that configuration \textit{LoRA-Warm}.
\chg{GRAFT is therefore the configuration used on the ten text tasks, where the
source and target backbones differ, and LoRA-Warm the one used on the five
multimodal tasks, where they coincide.}
\chg{For the text setting we additionally run a third configuration,
\textit{No bridge}, as an ablation: it transplants the retrieved $A$ and $B$
into Qwen2.5-VL unchanged. All other factors (retrieval, the selected anchor,
the data and the training budget) are kept fixed, so it differs from GRAFT
only in the bridge.}
\chg{The two comparisons differ. The multimodal one matches rank, scaling and
module coverage, so it isolates warm-starting on a single backbone rather than
the bridge; the text one is the cross-backbone test. Modality routing means
that a Qwen2.5 text adapter is never offered to an image-dependent task.}

\subsection{Evaluation}

All ten text tasks are in the closed label setting, and four of the five multimodal tasks are as well; only OCRVQA has an open answer space and is thus measured by exact match. All tasks are reported with Accuracy, with the change in macro-F1 alongside it. Both task types use the same epoch protocol: both configurations are trained for eight epochs, and each is evaluated at its own best-validation epoch on a 200-instance split disjoint from the test set, so the baseline has the same opportunity to select its strongest checkpoint as the warm-started one. Every reported cell is the mean over all four orders; the per-order values are given in Table~\ref{tab:app-perorder}. \chg{Text and multimodal means are reported separately, with a pooled mean over all fifteen tasks.} Each task is served by its own adapter, so we report per-task performance directly.

\subsection{Implementation Details}

Only LoRA parameters are updated; the Qwen2.5-VL backbone remains frozen.
Both configurations receive the same eight-epoch budget, with checkpoint selection as
described above. This measures adaptation under a finite training budget,
rather than assuming convergence, and matches the intended setting of
inexpensive adaptation as tasks arrive.


\section{Experimental Results}

Table~\ref{tab:results} reports both task types. For text tasks, each adapter is
initialised from a Qwen2.5 LoRA re-expressed through the activation-space
bridge; mean accuracy rises from $82.2$ to $83.3$, a gain of $1.17$ points
\chg{(Wilcoxon signed-rank, $p = 0.02$; Table~\ref{tab:sig})}.
For multimodal tasks, the retrieved adapters already work with Qwen2.5-VL and
are used directly; mean accuracy rises from $76.2$ to $78.3$, a gain of $2.13$
points.
\chg{Table~\ref{tab:ablation} isolates the bridge's share: transplanting the
retrieved adapter without it reaches $83.00$ against GRAFT's $83.66$ on
Order~A, a bridge contribution of $0.66$ points.}

\begin{table*}[!t]
\centering
\setlength{\tabcolsep}{6pt}
\begin{tabular}{@{}llcccc@{}}
\toprule
Task & Type & LoRA-Random & Warm-start & $\Delta$Acc & $\Delta$F1 \\
\midrule
ANLI Entailment & Text & 72.9 & \textbf{73.2} & $0.25$ & $0.41$ \\
Sentiment140 Polarity & Text & 88.2 & \textbf{89.7} & $1.45$ & $1.45$ \\
Sentiment140 Answer & Text & \textbf{91.2} & 90.3 & $-0.95$ & $-0.94$ \\
DVD Polarity & Text & 95.4 & \textbf{96.4} & $1.05$ & $1.05$ \\
Amazon Rating & Text & 62.8 & \textbf{64.5} & $1.70$ & $1.80$ \\
Amazon Food Rating & Text & 72.4 & \textbf{73.0} & $0.55$ & $0.95$ \\
Emotion & Text & 90.8 & \textbf{92.6} & $1.80$ & $1.77$ \\
MNLI Neutral & Text & 93.6 & 93.6 & $0.00$ & $-0.03$ \\
TellMeWhy & Text & 79.0 & \textbf{83.6} & $4.60$ & $2.76$ \\
DoQA Movies & Text & 75.5 & \textbf{76.7} & $1.25$ & $0.59$ \\
\cmidrule(lr){1-6}
Text mean & Text & 82.2 & \textbf{83.3} & $1.17$ & $0.98$ \\
\midrule
UC Merced Aerial & Multimodal & 70.6 & \textbf{73.0} & $2.40$ & $3.36$ \\
Dog Breed & Multimodal & 97.9 & \textbf{98.1} & $0.20$ & $0.18$ \\
RVL-CDIP & Multimodal & 66.6 & \textbf{67.5} & $0.90$ & $0.65$ \\
Waste Types & Multimodal & 84.0 & \textbf{85.2} & $1.25$ & $0.90$ \\
OCRVQA & Multimodal & 61.8 & \textbf{67.7} & $5.90$ & $-2.72$ \\
\cmidrule(lr){1-6}
Multimodal mean & Multimodal & 76.2 & \textbf{78.3} & $2.13$ & $0.47$ \\
\midrule
Overall mean & All & 80.2 & \textbf{81.7} & $1.49$ & $0.81$ \\
\bottomrule
\end{tabular}
\caption{Warm-start results across text and multimodal tasks. Text tasks use
GRAFT: the top-ranked retrieved Qwen2.5-7B adapter is re-expressed for
Qwen2.5-VL-7B through the bridge; multimodal tasks use the top-ranked
Qwen2.5-VL-7B adapter directly. No adapters are merged. Both configurations train for
eight epochs and are scored at their own best-validation epoch on 500 test
items per task. Accuracy is a percentage; $\Delta$Acc and $\Delta$F1 are
absolute differences in points, from unrounded scores. Every value is the mean
over four task orders, text and multimodal means computed separately; per-order
values are in Table~\ref{tab:app-perorder}. Class counts and training-set sizes
are in the appendix.}
\label{tab:results}
\end{table*}

\begin{table}[t]
\centering
\begin{minipage}[t]{0.42\linewidth}
\centering
\ifchg\color{red}\fi
\small
\setlength{\tabcolsep}{3pt}
\begin{tabular}{@{}lc@{}}
\toprule
Initialisation & Acc. \\
\midrule
No bridge & $83.00$ \\
GRAFT & $\mathbf{83.66}$ \\
\bottomrule
\end{tabular}
\caption{Bridge ablation. Mean accuracy on the ten text tasks of Order~A; the
two configurations differ only in the bridge.}
\label{tab:ablation}
\end{minipage}\hfill
\begin{minipage}[t]{0.56\linewidth}
\centering
\ifchg\color{red}\fi
\small
\setlength{\tabcolsep}{3pt}
\begin{tabular}{@{}ccc@{}}
\toprule
$\Delta$Acc & $95\%$ CI & $p$ \\
\midrule
$1.17$ & $[0.11, 2.23]$ & $\mathbf{0.02}$ \\
\multicolumn{3}{c}{} \\
\bottomrule
\end{tabular}
\caption{Gain over LoRA-Random on the ten text tasks, averaged over the four
orders, with an exact Wilcoxon signed-rank test and a $t$ interval on the mean
(Appendix~\ref{app:significance}).}
\label{tab:sig}
\end{minipage}
\end{table}


\section{Discussion}

Across the reported tasks, retrieved initialisation improves average accuracy
by $1.17$ points on text and $2.13$ on multimodal tasks. The gains are uneven:
TellMeWhy and OCRVQA benefit substantially, while Dog Breed is already close
to the baseline's ceiling. The practical value of reuse therefore depends on
both the retrieved capability and the remaining headroom under the training
budget, rather than on the mere availability of a pretrained adapter. \chg{Two separate points follow. Table~\ref{tab:ablation} shows
GRAFT exceeding the ablated \textit{No bridge} configuration in mean accuracy,
$83.66$ against $83.00$, so the bridge adds to warm-starting on average. And
the text gain over LoRA-Random in Table~\ref{tab:sig} is significant:
$1.17$ points, $95\%$ interval $[0.11, 2.23]$, $p = 0.02$.}

Sentiment140 Answer is the only task with a negative mean accuracy delta,
falling by $0.95$ points. Sentiment140 Polarity uses the same underlying tweet
corpus and retrieves the same source adapter, yet gains $1.45$ points on
average. The contrast suggests that transfer depends on task formulation:
direct polarity prediction benefits, while the yes/no task that verifies a
supplied sentiment does not. LoRA-Random already reaches $91.2\%$ on this
task, leaving limited headroom within the eight-epoch budget. GRAFT performs
best on Sentiment140 Answer under Order D (Table~\ref{tab:app-perorder}), the
only schedule that changes its position within the text-task sequence
(Appendix~\ref{sec:orders}); this is the strongest result observed within
that budget and suggests that the effectiveness of the warm start is also
sensitive to where the task occurs in the sequence. MNLI Neutral is
another boundary case. LoRA-Random already reaches $93.55\%$, leaving limited
headroom; the two configurations tie in mean accuracy, while macro-F1 differs by only
$-0.03$ points.

Order D changes the sequence within each modality (Appendix~\ref{sec:orders})
and has the largest number of negative cells (Table~\ref{tab:app-perorder}).
Amazon Rating and ANLI also change sign between orders, whereas TellMeWhy and
UC Merced Aerial remain positive in all four. Orders A--C preserve the same within-modality
sequence (Appendix~\ref{sec:orders}), so their interleaving does not expose a task to a different set of
earlier same-modality adapters. The order comparison should accordingly be
read as a stability check on the full pipeline rather than a causal estimate
of which schedule is best.

These comparisons include numerical variation as well as changes in library
state. Each configuration selects its checkpoint using validation accuracy, so small
training differences can change the reported epoch.

Accuracy and macro-F1 never reverse direction on the closed-label tasks;
MNLI Neutral's accuracy tie accompanies only a $0.03$-point F1 loss. OCRVQA
is the exception: its accuracy gain accompanies a macro-F1 loss, reflecting
the different weights the metrics give to frequent answers and rare answer
strings. Its benefit should therefore be read as improved exact-match
accuracy, rather than a uniform improvement across answer types.

These results support adapting existing Qwen2.5 LoRAs on Qwen2.5-VL,
without distillation or specially designed transferable adapters
\cite{translora,lorax}. Bridge folding removes inference overhead, while
separate task adapters prevent later training from overwriting earlier
parameters \cite{olora}.

Shared calibration and task-specific adaptation serve different purposes.
Calibration estimates a reusable relationship between the two backbones;
task supervision determines which parts of a retrieved capability are useful
for the new objective. Separate adapters retain task specialisations, while
folding lets them share one target backbone. Adding a task therefore requires
another lightweight adapter without retaining another source model for inference.

Retrieval remains separated by modality, and evaluation covers one backbone
pair. \chg{Qwen2.5-VL is built on Qwen2.5, so the two share a tokenizer and their
language modules correspond one to one. The bridge needs both, because paired
activations must come from the same token positions in matching modules. A
different model family, such as Llama, satisfies neither by default: a
tokenization alignment and a module correspondence would have to be established
first, which we do not study.} Ranking across both libraries
and transfer between less closely related models remain open questions.
\chg{The tasks were also chosen for capabilities the libraries already cover,
and we vary task order rather than the random seed.}


\section{Conclusion}
We presented GRAFT, a framework for reusing pretrained LoRA adapters across related language and multimodal backbones. GRAFT uses a learned linear bridge to map a retrieved source adapter into the target backbone before further adaptation. Experiments on ten text tasks and five multimodal tasks show that GRAFT consistently outperforms adaptation from randomly initialized LoRA adapters across most evaluated tasks. These results demonstrate the effectiveness of GRAFT's cross-backbone LoRA reuse mechanism for multimodal continual learning.


\bibliography{custom}

\clearpage
\appendix
\renewcommand{\thesection}{\Alph{section}}
\setcounter{section}{0}

\section*{Appendix}


\section{Dataset Details}

\paragraph{Text tasks.}
All text tasks are drawn from Super-NaturalInstructions \cite{sni} and are
fetched by task identifier at build time. For each task we cap training data at
3{,}000 instances, giving training sets of roughly 1{,}200 to 3{,}000 instances,
evaluate on 500 held-out instances, and hold out a further 200 instances for
epoch selection. Table~\ref{tab:app-text} lists the identifier for each task. ANLI pools the
three round-wise SNI entailment tasks into a single task, since the rounds
share a label space and differ only in the adversarial round that produced
them. The two Sentiment140 tasks share their tweet corpus but pose different
outputs: one predicts polarity directly, while the other answers yes or no to
a proposed polarity. Amazon Rating and Amazon Food Rating are two five-point
rating tasks over different Amazon review collections; the remaining tasks
draw on distinct source corpora.

\begin{table}[t]
\centering
\scriptsize
\setlength{\tabcolsep}{8pt}
\begin{tabular}{@{}l p{4.6cm}@{}}
\toprule
Task & Super-NaturalInstructions identifier \\
\midrule
ANLI Entailment & \texttt{task1385\_anli\_r1\_entailment},
                  \texttt{task1386\_anli\_r2\_entailment},
                  \texttt{task1387\_anli\_r3\_entailment} \\
\addlinespace[2pt]
Amazon Rating & \texttt{task1310\_amazonreview\_rating\_classification} \\
TellMeWhy & \texttt{task290\_tellmewhy\_question\_answerability} \\
Sentiment140 Polarity & \texttt{task195\_sentiment140\_classification} \\
Sentiment140 Answer & \texttt{task196\_sentiment140\_answer\_generation} \\
Amazon Food Rating & \texttt{task588\_amazonfood\_rating\_classification} \\
Emotion & \texttt{task875\_emotion\_classification} \\
MNLI Neutral & \texttt{task201\_mnli\_neutral\_classification} \\
DVD Polarity & \texttt{task477\_cls\_english\_dvd\_classification} \\
DoQA Movies & \texttt{task1442\_doqa\_movies\_isanswerable} \\
\bottomrule
\end{tabular}
\caption{The ten Super-NaturalInstructions text tasks. All are evaluated with
Accuracy. ANLI pools the three round-wise entailment tasks.}
\label{tab:app-text}
\end{table}

\paragraph{Multimodal tasks.}
Four of the five multimodal tasks are built from publicly available image
classification datasets, listed in Table~\ref{tab:app-vision}; the fifth,
OCRVQA, is an OCR-based visual question answering task with an open answer
space rather than a fixed class list. As with the text
tasks, training data is capped at 3{,}000 instances and evaluation uses 500
held-out instances. For the four classification tasks, each prompt presents
the candidate class list, which is derived from the training split alone so
that no evaluation-time label information enters the prompt.

\begin{table}[t]
\centering
\scriptsize
\setlength{\tabcolsep}{10pt}
\begin{tabular}{@{}llrr@{}}
\toprule
Task & Source dataset & $|C|$ & $|$train$|$ \\
\midrule
UC Merced Aerial   & UC Merced Land Use       & 21 & 1{,}400 \\
Dog Breed          & animime-dog-breeds       & 10 &    635 \\
RVL-CDIP           & RVL-CDIP (subset)        & 16 & 1{,}860 \\
Waste Types        & waste-classification     & 18 & 1{,}276 \\
OCRVQA             & OCR-VQA                  & n/a & 3{,}000 \\
\bottomrule
\end{tabular}
\caption{The five multimodal tasks. $|C|$ is the number of classes; OCRVQA
has an open answer space and no fixed class list.}
\label{tab:app-vision}
\end{table}


\section{Adapter Library Composition}

The multimodal pool comprises 22 adapters collected for
Qwen2.5-VL-7B-Instruct, of which 20 are compatible with the generative
backbone; the two exclusions are retrieval adapters whose multi-vector
projection head has no counterpart in the generative model. The pool spans
visual question answering, OCR and document conversion, aerial and driving
scene understanding, visual and narrative reasoning, and multimodal quality
evaluation.

\chg{Descriptions and keywords for all library adapters were generated in a
single offline pass with Claude Opus 5, prompted with each adapter's
HuggingFace model card. The schema is fixed: a one-paragraph description naming
the task, the domain, the inputs, the outputs and four capabilities, followed
by six to ten keywords. The generated text is used only to rank candidates
against an incoming task definition; it never enters training or evaluation, so
an inaccurate description can only cause a less suitable adapter to be
retrieved, which works against the warm-started configuration rather than for
it. We did not manually verify the generated fields against each adapter's
configuration.}

The text pool comprises 28 LoRAs trained on Qwen2.5-7B-Instruct, covering
natural language inference, commonsense and multi-step reasoning,
summarization, question answering, information extraction, sentiment and
emotion classification, and domain-specific knowledge, principally medical.

The collected adapters are heterogeneous in configuration. Across the text
pool, ranks span $r \in \{8, 16, 24, 32, 64, 128\}$ with $r=16$ most common,
LoRA coefficients $\alpha$ range from 8 to 128, dropout takes values in
$\{0, 0.05, 0.075\}$, and target-module selections cover two, four, or all
seven linear projections.


\section{Implementation Details}

Adapters are optimized with AdamW at a learning rate of 3e-5 and no weight
decay, with a per-step batch size of one and gradient accumulation over four
steps. Gradient norms are clipped to 1.0 and sequences longer than 1{,}024
tokens are skipped. All runs use seed 42 on a single NVIDIA A100 SXM4 40GB GPU.

The adapter configurations are as follows:
\begin{itemize}
    \item \textit{Text, LoRA-Random:} rank $r=16$, LoRA coefficient
    $\alpha=32$ (update scaling $s=\alpha/r=2$), and all seven projection
    types: \texttt{q\_proj}, \texttt{k\_proj}, \texttt{v\_proj},
    \texttt{o\_proj}, \texttt{gate\_proj}, \texttt{up\_proj}, and
    \texttt{down\_proj}. $A$ is Kaiming-uniform and $B$ is zero. Keeping
    this configuration fixed across text tasks provides a consistent reference
    for adaptation from scratch, without making the baseline's architecture
    depend on which adapter retrieval selects.
    \item \textit{Text, GRAFT:} the retrieved adapter supplies the initial
    matrices, rank, scaling, and module set; the activation bridge is held fixed
    during training. Retaining this configuration allows reuse of existing
    adapters without an additional rank-conversion step or changes to their
    module coverage, either of which would modify the pretrained structure
    being reused. The text comparison therefore evaluates the complete reuse
    procedure against a fixed baseline for adaptation from scratch.
    \chg{Capacity is not matched, but the mismatch does not favour GRAFT. Rank
    coincides at $r=16$ on nine of the ten tasks and the update scaling is
    $s=2$ in both configurations; ANLI is the only task whose retrieved adapter is larger
    ($r=24$). The retrieved adapters instead cover fewer module types: on seven
    tasks they adapt only the four attention projections, giving $10.1$M
    adapter parameters against the baseline's $40.4$M. Those seven tasks also
    show the largest mean gain, $+2.06$ points, against $+0.70$ on the two
    tasks where coverage coincides and $+1.00$ on ANLI, so the improvement is
    not explained by additional capacity. The GRAFT and no-bridge rows of
    Table~\ref{tab:ablation} share rank, scaling, module set and anchor
    exactly, so that contrast isolates the bridge at matched capacity.}
    \item \textit{Multimodal, LoRA-Warm:} the retrieved adapter's pretrained
    matrices, rank, scaling, and module set are retained. It is used directly
    because it was trained on the same Qwen2.5-VL backbone.
    \item \textit{Multimodal, LoRA-Random:} the retrieved adapter's matrix
    shapes, module set, and scaling are retained, but $A$ is reinitialised with
    Kaiming-uniform values and $B$ is set to zero. This matches capacity within
    each task so that the configurations differ only in the initial values of their LoRA
    $A$ and $B$ matrices.
\end{itemize}

For retrieval, candidates are ranked by cosine similarity between the encoded
task definition and each adapter's encoded metadata, and the top-ranked
adapter is used; no loss-based reranking is performed for any reported result,
and the ranking is deterministic given the task description and current
library contents.

\chg{Routing is rule-based and deterministic, with a fixed order of authority.
An explicit modality declaration in the task file takes precedence; a
declaration in the task registry is used next; failing both, up to 64 of the
task's own samples are probed and the task is routed to the multimodal pool if
any sample carries image content. A task registered under both modalities that
declares neither is rejected rather than guessed. No learned classifier is
involved, and the rule applied and the evidence for it are recorded for every
task. The text library holds text adapters only, either pretrained Qwen2.5
LoRAs that take the bridge or Qwen2.5-VL text LoRAs warm-started on earlier
text tasks, so a text task is never served an adapter from the multimodal
library.}

\chg{A returned adapter is registered with keywords naming the new task, and
is then ranked by the same similarity score as the pretrained LoRAs.}

\chg{\subsection{Significance of the Text Result}}
\label{app:significance}

\chg{Significance is assessed over tasks rather than over test items, because
the task is the replicated unit: each task is served by its own adapter and
scored on its own test split. Applied to the per-task $\Delta$Acc values of
Table~\ref{tab:app-perorder}, an exact two-sided Wilcoxon signed-rank test over
the ten text tasks gives $p = 0.0195$ ($W^{+} = 42$, $W^{-} = 3$; MNLI Neutral
is tied at $0.00$ and is dropped). A $95\%$ interval on the mean gain, from the
$t$ distribution over the same ten values, is $[+0.11, +2.23]$, and is the
interval reported in Table~\ref{tab:sig}. Table~\ref{tab:ablation} reports
mean accuracy only; the difference between the two warm-started initialisations
is not tested, since one order gives a single paired value per task. All
reported numbers come from a single run of each configuration; significance is
therefore assessed across tasks rather than across independent training runs.}

\begin{table*}[t!]
\centering
\setlength{\tabcolsep}{3pt}
\begin{tabular}{@{}lccccc@{}}
\toprule
Task & A & B & C & D & Mean \\
\midrule
\multicolumn{6}{@{}l}{\textit{Text}} \\[1pt]
ANLI Entailment & $+1.00$ & $+1.80$ & $-0.20$ & $-1.60$ & $+0.25$ \\
Sentiment140 Polarity & $+2.00$ & $+2.00$ & $+2.00$ & $-0.20$ & $+1.45$ \\
Sentiment140 Answer & $-1.20$ & $-1.20$ & $-1.20$ & $-0.20$ & $-0.95$ \\
DVD Polarity & $+1.00$ & $+1.20$ & $+0.60$ & $+1.40$ & $+1.05$ \\
Amazon Rating & $+5.00$ & $-3.00$ & $+1.00$ & $+3.80$ & $+1.70$ \\
Amazon Food Rating & $+1.20$ & $+1.20$ & $+0.00$ & $-0.20$ & $+0.55$ \\
Emotion & $+1.40$ & $+1.40$ & $+1.40$ & $+3.00$ & $+1.80$ \\
MNLI Neutral & $+0.00$ & $+0.00$ & $-0.80$ & $+0.80$ & $+0.00$ \\
TellMeWhy & $+4.60$ & $+4.60$ & $+4.60$ & $+4.60$ & $+4.60$ \\
DoQA Movies & $+1.80$ & $+0.20$ & $+0.80$ & $+2.20$ & $+1.25$ \\
\midrule
\multicolumn{6}{@{}l}{\textit{Multimodal}} \\[1pt]
UC Merced Aerial & $+2.40$ & $+2.40$ & $+2.40$ & $+2.40$ & $+2.40$ \\
Dog Breed & $+0.00$ & $+0.20$ & $+0.80$ & $-0.20$ & $+0.20$ \\
RVL-CDIP & $+1.20$ & $+2.00$ & $+0.00$ & $+0.40$ & $+0.90$ \\
Waste Types & $+2.20$ & $+0.40$ & $+0.80$ & $+1.60$ & $+1.25$ \\
OCRVQA & $+6.00$ & $+6.00$ & $+6.00$ & $+5.60$ & $+5.90$ \\
\bottomrule
\end{tabular}
\caption{Change in accuracy, in points, under the four task orders listed in
Appendix~\ref{sec:orders}. Each order includes the same ten text and five
multimodal tasks. The Mean column averages A--D and matches $\Delta$Acc in
Table~\ref{tab:results}. Differences between orders include training and
checkpoint-selection variation and do not isolate an effect of library contents.}
\label{tab:app-perorder}
\end{table*}

\subsection{Bridge Calibration}
\label{app:bridge-calibration}
\chg{The calibration corpus is assembled round-robin from the training splits
of a fixed set of six text tasks -- SentimentTwitter, CommonsenseVal, ANLI,
PubMedQA, ExtremeSumm and EnDeTranslation -- held over from earlier
single-backbone experiments and unchanged for every run reported here,}
after shuffling each split with seed 0. Prompts
and responses are concatenated and truncated to 256 tokens. Identical token
IDs are fed to both frozen backbones. We fit the bridge on 461 sequences
(65,102 tokens) and reserve 51 further sequences for the held-out alignment
diagnostic. For each input space, the ridge coefficient $\lambda$ is
$10^{-3}$ times the mean diagonal of $X_T^{\top}X_T$, with that mean floored
at $10^{-6}$. The systems are solved in float32 and the transfer matrices
are cached in float16. Sharing a map among projections with the same input
yields 84 matrices across 28 language layers, covering 168 modules. The
\texttt{down\_proj} modules have no transfer matrix and use their LoRA
matrices without this transform. The cache is fitted once and shared across
all source adapters.
\chg{An activation-space map is fitted from paired activations, so a
calibration corpus is required by construction. Like the frozen backbones, the
map is estimated once before the sequence and is never updated by it: it
belongs to the system's setup rather than to task learning. Calibration uses
training splits only, with no test item, and the regression target is the
source backbone's activations rather than the response, so no task supervision
enters the map. One map serves every task and every adapter, so it cannot
favour a particular one. The task set was fixed before these
experiments and was not selected for them.}


\section{Task Orders}
\label{sec:orders}

The library grows as tasks arrive, so a task retrieves from a different pool
depending on where it falls in the sequence. To evaluate performance across
different task sequences, the benchmark is run in the four orders below, all fixed
before any was run. Multimodal tasks are shown in \textit{italics}.

Orders A, B and C share the same within-modality sequence and differ only
in how the two modalities interleave.

\paragraph{Order A - all text, then all multimodal.}
Emotion $\rightarrow$ Sentiment140 Polarity $\rightarrow$ MNLI Neutral $\rightarrow$ TellMeWhy $\rightarrow$ Amazon Rating $\rightarrow$ Amazon Food Rating $\rightarrow$ ANLI Entailment $\rightarrow$ DVD Polarity $\rightarrow$ Sentiment140 Answer $\rightarrow$ DoQA Movies $\rightarrow$ \textit{Dog Breed} $\rightarrow$ \textit{OCRVQA} $\rightarrow$ \textit{UC Merced Aerial} $\rightarrow$ \textit{Waste Types} $\rightarrow$ \textit{RVL-CDIP}

\paragraph{Order B - all multimodal, then all text.}
\textit{Dog Breed} $\rightarrow$ \textit{OCRVQA} $\rightarrow$ \textit{UC Merced Aerial} $\rightarrow$ \textit{Waste Types} $\rightarrow$ \textit{RVL-CDIP} $\rightarrow$ Emotion $\rightarrow$ Sentiment140 Polarity $\rightarrow$ MNLI Neutral $\rightarrow$ TellMeWhy $\rightarrow$ Amazon Rating $\rightarrow$ Amazon Food Rating $\rightarrow$ ANLI Entailment $\rightarrow$ DVD Polarity $\rightarrow$ Sentiment140 Answer $\rightarrow$ DoQA Movies

\paragraph{Order C - interleaved, two text to one multimodal.}
Emotion $\rightarrow$ Sentiment140 Polarity $\rightarrow$ \textit{Dog Breed} $\rightarrow$ MNLI Neutral $\rightarrow$ TellMeWhy $\rightarrow$ \textit{OCRVQA} $\rightarrow$ Amazon Rating $\rightarrow$ Amazon Food Rating $\rightarrow$ \textit{UC Merced Aerial} $\rightarrow$ ANLI Entailment $\rightarrow$ DVD Polarity $\rightarrow$ \textit{Waste Types} $\rightarrow$ Sentiment140 Answer $\rightarrow$ DoQA Movies $\rightarrow$ \textit{RVL-CDIP}

\paragraph{Order D - fully shuffled.}
This order shuffles all fifteen tasks together, changing both their
within-modality positions and how text and multimodal tasks mix. It tests
adaptation to a mixed task stream without modality blocks or a fixed
alternating pattern.

Amazon Food Rating $\rightarrow$ \textit{OCRVQA} $\rightarrow$ DVD Polarity $\rightarrow$ \textit{Dog Breed} $\rightarrow$ DoQA Movies $\rightarrow$ TellMeWhy $\rightarrow$ ANLI Entailment $\rightarrow$ \textit{RVL-CDIP} $\rightarrow$ Emotion $\rightarrow$ Sentiment140 Polarity $\rightarrow$ MNLI Neutral $\rightarrow$ \textit{Waste Types} $\rightarrow$ Amazon Rating $\rightarrow$ \textit{UC Merced Aerial} $\rightarrow$ Sentiment140 Answer


\section{Environment}
\label{sec:repro}

All experiments were run on NVIDIA A100-SXM4 40\,GB GPUs under PyTorch 2.11
(CUDA 12.8), Transformers 5.8.1 and Python 3.12. The seed, the
train/validation/test splits, the task orders, the retrieval ranking and the
initialisation of both configurations are fixed by the released setup.

Numerical results depend on the execution environment as well as on the
configuration. We use PyTorch's fused scaled dot-product attention, which is
considerably faster than the deterministic reference implementation at these
sequence lengths; the fused kernels accumulate partial results in an order that
is not fixed, and the resulting differences fall in the last bits of the backward
pass, so small differences are expected on different hardware, with
different kernel implementations, or under different library versions. Because
each configuration is scored at the epoch with the best validation accuracy, such a
difference can also change which epoch is selected. Every reported cell is the
mean over four task orders.

\end{document}